\documentclass[letterpaper]{article} % DO NOT CHANGE THIS
\usepackage{aaai23}  % DO NOT CHANGE THIS
\usepackage{times}  % DO NOT CHANGE THIS
\usepackage{helvet}  % DO NOT CHANGE THIS
\usepackage{courier}  % DO NOT CHANGE THIS
\usepackage[hyphens]{url}  % DO NOT CHANGE THIS
\usepackage{graphicx} % DO NOT CHANGE THIS
\usepackage{natbib}  % DO NOT CHANGE THIS AND DO NOT ADD ANY OPTIONS TO IT
\usepackage{caption} % DO NOT CHANGE THIS AND DO NOT ADD ANY OPTIONS TO IT
\usepackage{algorithm}
\usepackage{algorithmic}

\usepackage{newfloat}
\usepackage{listings}
\DeclareCaptionStyle{ruled}{labelfont=normalfont,labelsep=colon,strut=off} % DO NOT CHANGE THIS
\floatstyle{ruled}
\newfloat{listing}{tb}{lst}{}
\floatname{listing}{Listing}

\title{Challenges facing the explainability of age prediction models: 

case study for two modalities}
\author{
       Mikołaj Spytek${}^1$\,\,\,\, Weronika Hryniewska-Guzik${}^1$\, \\ Jarosław Żygierewicz${}^2$\,\,\,\, Jacek Rogala${}^2$\, \\  Przemysław Biecek${}^{1,2}$
}
\affiliations{
    ${}^1$  Warsaw University of Technology \\
    ${}^2$  University of Warsaw \\
}

\usepackage{bibentry}
\begin{document}

\maketitle

\begin{abstract}
The prediction of age is a challenging task with various practical applications in high-impact fields like the healthcare domain or criminology. 

Despite the growing number of models and their increasing performance, we still know little about how these models work. 
Numerous examples of failures of AI systems show that performance alone is insufficient, thus, new methods are needed to explore and explain the reasons for the model's predictions.

In this paper, we investigate the use of Explainable Artificial Intelligence (XAI) for age prediction focusing on two specific modalities, EEG signal and lung X-rays. We share predictive models for age to facilitate further research on new techniques to explain models for these modalities.

\end{abstract}


\section{Introduction}

Models for age prediction have many interesting applications. Age can be a direct target of a model, for example in age prediction for forensics to identify the age of the offender, as well as indirect, for example, as an input feature. The list of examples is extensive, just to name a few: age prediction is useful feature in cancer diagnostics \cite{cancer21}, analyses of brain ageing lead to relevant features for predicting neurological disorders, such as schizophrenia or bipolar disorder \cite{BrainAGE17} or cognitive impairment \cite{NeuroImage17}.

Age prediction can be performed on modalities that differ in terms of structure and format. Just to list some of these modalities: 2d images of a person's face (features such as wrinkles, crow's feet, and age spots can be used to predict a person's age \cite{CVPRWage15}), 1d voice recordings (features such as pitch, tone, and accent can be used to predict a person's age), series of nucleotides such as methylation data \cite{methylclock20}, multiple series as in EEG \cite{EEGage19, ColeBrain17} or single-channel images X-rays \cite{Thodberg2017}. 

This variety of modalities creates a major challenge when trying to explain predictive models. It is because the most popular methods of explainable AI are agnostic to the structure of the model but they are not agnostic to the modality of the data \cite{Holzinger2022, ema21}. If they are designed for tabular, image, or text data, it is difficult to transfer them to EEG signals or grayscale images.

To address this problem, we created and shared a collection of benchmark models for two modalities that currently do not have representations suitable for explanations.

\section{Materials and methods}

A series of models has been trained for the age prediction problem and are available on GitHub repository \url{https://github.com/pbiecek/challenges-xai-aging-aaai23}. 

The models vary in structure, ranging from linear regression models, which are  interpretable by design, to tree based boosted models and multi-layer neural networks. The following sections summarise the data on which these models were trained.

\subsection{Age prediction with EEG data}

The first use case concerns predicting the metrical age of patients based on EEG data. The dataset containing 20,365 observations of patients aging from 16 to 70 has been gathered over the past 15 years across 32 hospitals and medical institutes in Poland. All recordings contain measurements from 19 electrodes placed according to the international 10-20 system \cite{system10-20} with sampling rates of 250, 256, or 500 Hz, depending on the used hardware. For the purposes of this analysis, only the data from healthy patients was used.

The raw recording data has been preprocessed to extract 1,653 features for each observation. The first 1,368 features represent each of 8 frequency bands multiplied by the 171 values from the lower triangle of the coherence matrix. The remaining features correspond to the normalized band power from 15 frequency bands for each of the 19 channels. 

\begin{figure*}
\centering
    \includegraphics[width=0.35\textwidth]{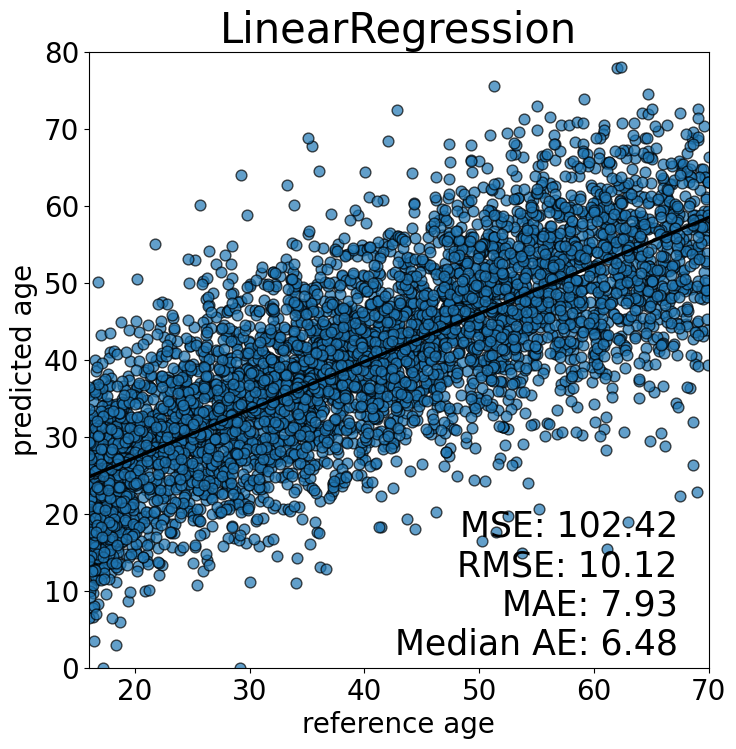}
    \includegraphics[width=0.35\textwidth]{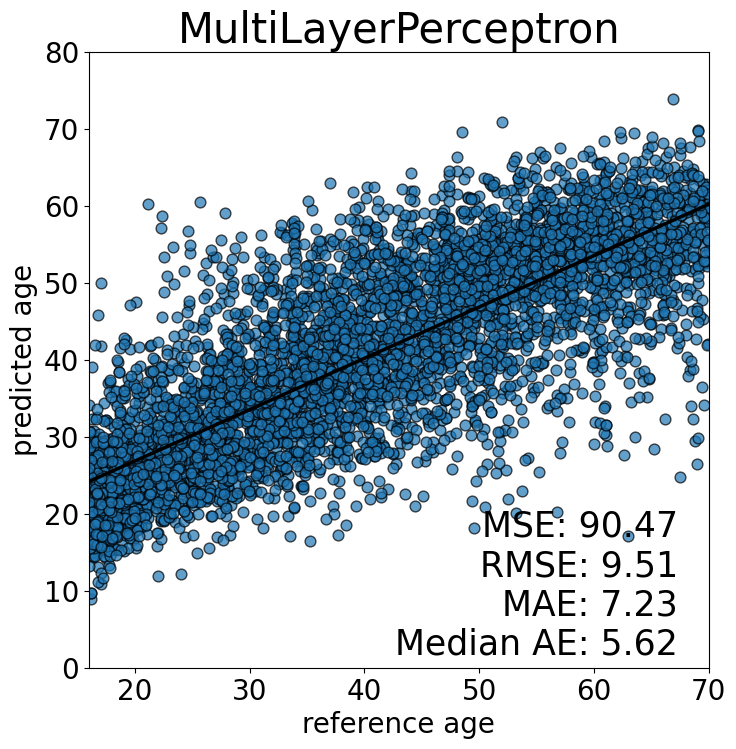}
\caption{
Diagnostic plots for developed age prediction models for EEG data. A) predictions from a Linear Regression model, B) predictions from a Multi-Layer Perceptron}
\label{fig:eeg-models}
%\end{figure*}

\vspace{0.5cm}

%\begin{figure*}[h]
\centering
    \includegraphics[width=0.35\textwidth]{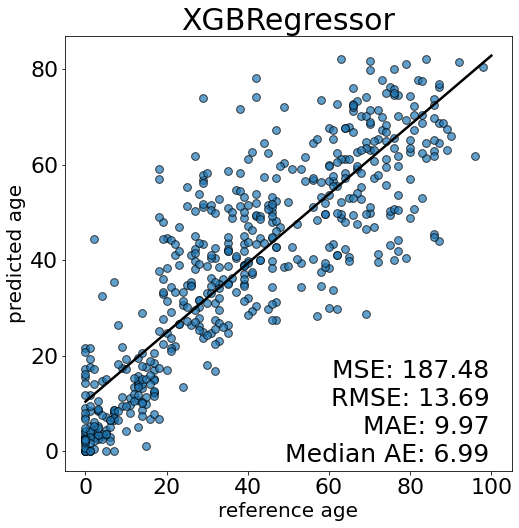}
    \includegraphics[width=0.35\textwidth]{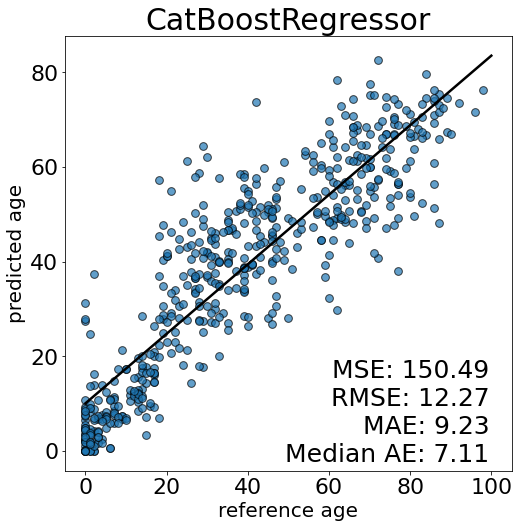}
\caption{
Diagnostic plots for developed age prediction models for X-ray data. 
A) predictions from XGBoost Regression model, B) predictions from Cat Boost Regression model}
\label{fig:xray-models}
\end{figure*}

Using the extracted features two machine learning models were fitted, with the task of predicting the age of the patient -- a Linear Regression model and a Multi-Layer Perceptron. Predictions on a holdout test set of 5,091 observations are presented in Figure~\ref{fig:eeg-models}. The linear regression achieves a Mean Absolute Error of 7.93 years, whereas the MLP network scores 7.23 in this metric. %Interestingly, for both models the predictions flatten out for older patients. This can mean that starting from a certain age, the recorded brain activity is a worse predictor of the patient's age or that the differences in brain activity of older people are less pronounced.

\subsection{Age prediction with x-ray lung data}

The second use case is based on a new B2000 dataset with X-rays images. The data contains cases from two hospitals in Poland it is composed of adults and children X-rays. Each image is annotated with labels from 6 different classes: normal, pneumothorax, airspace opacification, fluid, cardiomegaly, and mass/nodule. As mentioned in~\cite{checklist}, images of children and adults should not be mixed in one class in the standard classification task. However, embeddings are representations of images, so we decided to use one of the few-shot learning methods, namely siamese networks. Few-shot learning helps to deal with a problem when a few examples in each class are present.

We trained the Siamese network with 2,158 images for 20 epochs and a learning rate 0.0005. The backbone was EfficientNet-B0 of input size 224x224. The baseline vector (anchor vector) is compared with a positive and negative one, and the resulting values are passed to the triplet loss. The generated embeddings were vectors of length 512.

The XGBoost regressor with squared loss and the CatBoost regressor were trained on embeddings retrieved from the model of the lowest validation loss to predict the age of a patient. After training, we clipped the age values below zero to zero. Obtained results are presented in Figure \ref{fig:xray-models}.

\section{Conclusions}

To facilitate the development of XAI techniques for regression models that are needed to explain age prediction models in the healthcare domain, it is crucial to share with the XAI community specific cases on which such techniques can be benchmarked and validated.

In this work, we have developed and shared age prediction models operating on two modalities for which there are currently no good explanatory techniques, that is, multidimensional series in the EEG signals and grayscale X-ray images. 

We hope that the availability of these models and validation samples will help in the development of new explanatory techniques for age prediction.

\vskip 0.2in
\bibliography{sample}

\end{document}